\pdfoutput=1

\documentclass[11pt,a4paper]{article}

\usepackage[]{acl}

\usepackage{times}
\usepackage{latexsym}

\usepackage[T1]{fontenc}

\usepackage[utf8]{inputenc}

\usepackage{microtype}
\usepackage{booktabs}
\usepackage{times}
\usepackage{latexsym}

\usepackage{url}

\usepackage{caption}
\usepackage{subfigure}
\usepackage{booktabs}
\usepackage{latexsym}
\usepackage{amsmath}
\usepackage{xspace}
\usepackage{amssymb}
\usepackage{graphicx}
\usepackage{multirow}
\usepackage{enumitem}
\usepackage[ruled,vlined]{algorithm2e}
\usepackage{algorithmic}
\usepackage{tablefootnote}
\usepackage{ulem}

\SetKwInput{KwInput}{Input}                
\SetKwInput{KwOutput}{Output}              

\DeclareMathOperator{\softmax}{softmax}
\DeclareMathOperator{\ReLU}{ReLU}
\DeclareMathOperator{\crossentropy}{cross-entropy}

\newcommand{\ie}{\textit{i.e.,}\xspace}
\newcommand{\eg}{\textit{e.g.,}\xspace}
\newcommand{\etal}{\textit{et al.}\xspace}
\newcommand{\paratitle}[1]{\vspace{0.8ex}\noindent \textbf{#1}}

\title{A Dual-Channel Framework for Sarcasm Recognition by Detecting Sentiment Conflict}

\author{Yiyi Liu$^{\text{*}1,2}$, Yequan Wang$^{\text{*}3}$, Aixin Sun$^4$, Xuying Meng$^5$, Jing Li$^6$, Jiafeng Guo$^{1,2}$\\
        $^1$CAS Key Lab of Network Data Science and Technology, Institute of Computing Technology,\\ Chinese Academy of Sciences, Beijing, China\\
        $^2$University of Chinese Academy of Sciences, Beijing, China\\
        $^3$Beijing Academy of Artificial Intelligence, Beijing, China\\
        $^4$School of Computer Science and Engineering, Nanyang Technological University, Singapore \\
        $^5$Institute of Computing Technology, Chinese Academy of Sciences, Beijing, China\\
        $^6$Inception Institute of Artificial Intelligence, Abu Dhabi, United Arab Emirates \\
        \tt tshwangyequan@gmail.com,\tt axsun@ntu.edu.sg,\tt jingli.phd@hotmail.com\\
        \tt \{liuyiyi17s,mengxuying,guojiafeng\}@ict.ac.cn
        }

\begin{document}
\maketitle

\renewcommand{\thefootnote}{\fnsymbol{footnote}}
\footnotetext[0]{*Indicates equal contribution}
\renewcommand{\thefootnote}{\arabic{footnote}}

\begin{abstract}

Sarcasm employs ambivalence, where one says something positive but actually means negative, and vice versa. 
The essence of sarcasm, which is also a sufficient and necessary condition, is the conflict between literal and implied sentiments expressed in one sentence. However, it is difficult to recognize such sentiment conflict because the sentiments are mixed or even implicit.
As a result, the recognition of sophisticated and obscure sentiment brings in a great challenge to sarcasm detection. 
In this paper, we propose a Dual-Channel Framework by modeling both literal and implied sentiments separately. Based on this dual-channel framework, we design the Dual-Channel Network~(DC-Net) to recognize sentiment conflict.
Experiments on political debates (\ie IAC-V1 and IAC-V2) and Twitter datasets show that our proposed DC-Net achieves state-of-the-art performance on sarcasm recognition.  Our code is released to support research\footnote{\url{https://github.com/yiyi-ict/dual-channel-for-sarcasm}}.

\end{abstract}

\section{Introduction}
\label{sec:intro}

Sarcasm is a complicated linguistic phenomenon. Intuitively, it means that one says something positive on surface form, while he/she actually expresses negative, vice versa~\cite{DBLP:series/synthesis/2012Liu,Merrison+2008+331+334}. 
Take the sentence ``\normalem{\emph{\uline{Final exam is the best gift on my birthday}}}'' as an example, the literal sentiment on surface is \textit{positive}, which is reflected by the explicit sentiment words, \ie ``\textit{best gift}''. 
However, the factual part of the text (\ie ``\textit{final exam happens on birthday}'') implies that the sentiment expressed is \textit{negative}.
This example suggests that it is the sentiment conflict that causes sarcasm linguistically.

However, modeling this linguistic nature of sarcasm is a great challenge due to the difficulty of digging  sentiment conflict between the literal and the implied meanings.
We know that non-sarcastic texts do not contain implied meaning, so the literal sentiment is consistent with the actual sentiment. 
But for sarcastic text, there is more than one meaning that coexists in one sentence. The literal meaning and the implied meaning are reflected in different sub-sentences. Even more challenging, sentiments behind the two meanings are mixed or even implicit.

Many existing studies adopt generic classification models for sarcasm recognition~\cite{DBLP:conf/sigir/LouL0HDX21,DBLP:conf/wassa/GhoshV16}. 
However, these methods directly model the entire sentence without considering the contradictory meanings behind sarcastic texts. 
There are also studies using contrast patterns (\eg phrase pair and word pair) as indicators to detect sarcasm, which is approaching the linguistic essence of sarcasm. 
\citet{riloff2013sarcasm,joshi2015harnessing} detect contrast or incongruity patterns, \ie the co-occurrence of positive sentiment phrases and negative situational phrases.
\citet{DBLP:conf/acl/SuTHL18,DBLP:conf/www/XiongZZY19} use attention mechanism to measure the sentiment conflict between word pairs in sarcastic texts. 
However, these methods emphasize too much on the explicit sentiment conflict on surface form~(\ie word/phrase level), which mainly reflect the literal meaning. 
As a result, the factual text is underestimated, which expresses the implied sentiment. 

\begin{figure}
    \centering
    \includegraphics[width=.99\columnwidth]{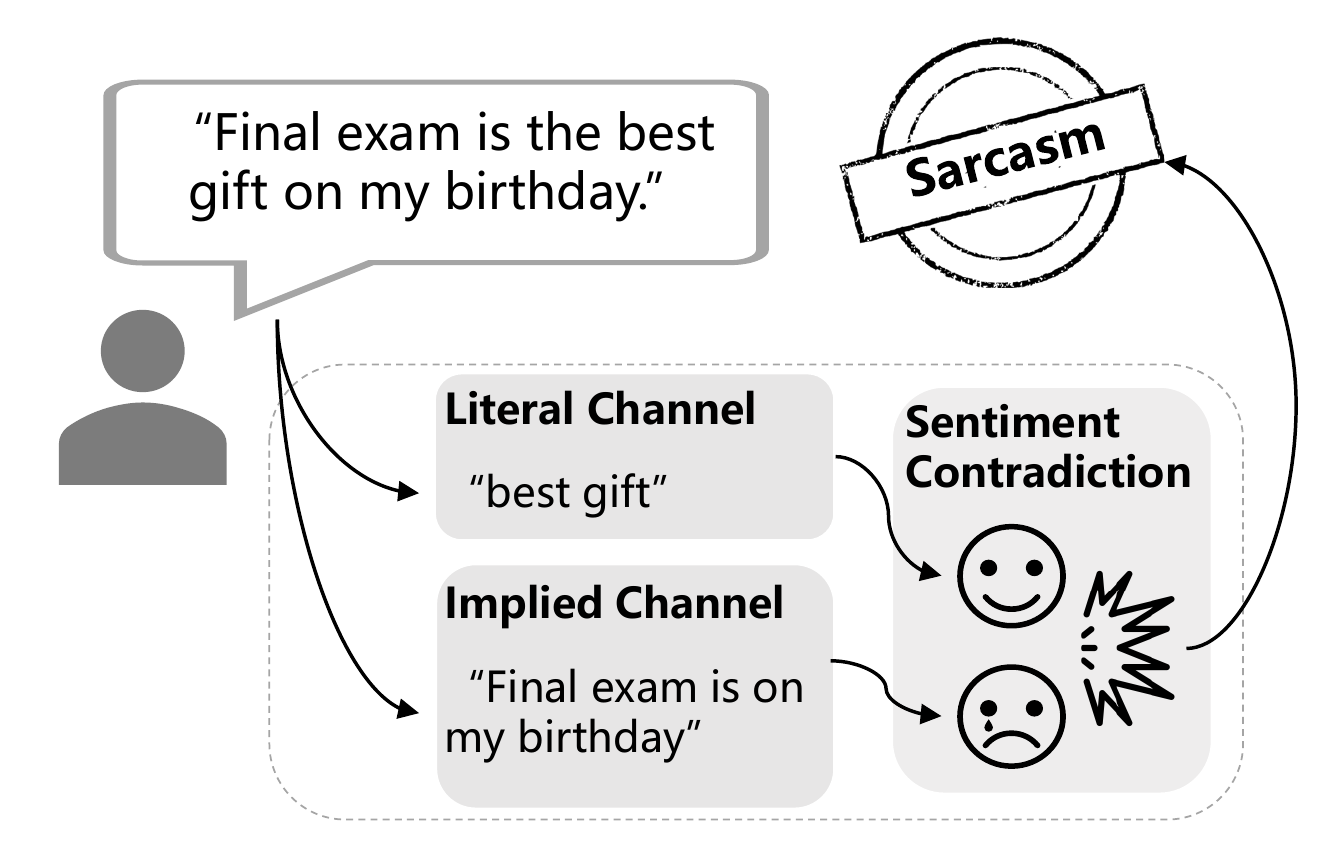}
    \caption{The Dual-Channel Framework for sarcasm recognition.}
    \label{fig:frame}
\end{figure}

\paratitle{Dual-Channel Framework.} 
In this paper, we propose a dual-channel framework to model the literal sentiment and the implied sentiment simultaneously. This allows us to leverage the conflict between the two channels in a comprehensive way.
Figure~\ref{fig:frame} depicts the proposed dual-channel framework. In this framework, literal channel and implied channel are used to detect the surface and the hidden meanings separately. Once sentiment conflict is detected, we could determine the existence of sarcasm.
The design of our dual-channel framework balances the effect of literal and implied inputs and avoids focusing too much on either one channel while ignoring the other. 
Our framework covers existing sarcasm patterns, and could be further enhanced to detect more sentiment conflict patterns.

Based on this framework, we develop the Dual-Channel Network~(DC-Net) to detect sarcasm. DC-Net contains four modules: decomposer, literal channel, implied channel, and analyzer.
In general, sentiment words directly reflect the surface sentiment, while the text without sentiment words reflects the implied sentiment. Hence, we split the sentiment words of input text to literal channel, and the remaining words to implied channel by decomposer. 
Then we use the literal channel to model surface meaning, and the implied channel to model hidden meaning. Lastly, we use analyzer to recognize the conflict.
Experiments on three benchmark datasets (\ie, IAC-V1, IAC-V2 and Tweets) show that our proposed DC-Net model achieves state-of-the-art  performance.

The main contributions of this paper are twofold. First, to the best of our knowledge, the dual-channel framework is the first attempt to explicitly separate literal meaning and implied meaning to recognize sarcasm by detecting sentiment conflict.
Second,  experiments conducted on benchmark datasets (\ie IAC-V1/V2 and Tweets) show that DC-Net achieves state-of-the-art performance.

\section{Related Work}
\label{sec:relatedwork}

Prior methods of sarcasm recognition can be divided into traditional models and neural models. 
There are also methods considering context information, \eg posting history~\cite{DBLP:conf/coling/HazarikaPGCZM18,DBLP:conf/coling/ZhangZF16}, and user profile~\cite{DBLP:conf/coling/PoriaCHV16,DBLP:conf/emnlp/KolchinskiP18}.
However, such context may not be always available.

\subsection{Traditional Models}

Most traditional approaches adopt machine learning methods such as SVM with manually crafted rules or feature engineering.
The features include sentiment lexicons~\cite{DBLP:conf/acl/Gonzalez-IbanezMW11,DBLP:conf/cicling/PatraM0RB16}, pragmatic features (\ie emoticons~\cite{gonzalez2011identifying}, capitalization, punctuations~\cite{joshi2015harnessing}), and pattern-based features~\cite{riloff2013sarcasm}~\etal. \citet{DBLP:journals/coling/HeeLH18} utilize common sense to assist sarcasm detection on Twitter.
Accordingly, the accuracy of sarcasm recognition highly depends on the quality of features.

Rewriting key parts of a sentence manually is an expensive but effective method.
~\citet{DBLP:conf/emnlp/GhoshGM15} believe that sarcasm involves a figurative meaning which is usually the opposite of literal meaning. They reframe sarcasm recognition as a literal/sarcastic word sense disambiguation problem. Then they paraphrase sarcastic texts manually to obtain target words that cause sarcastic disambiguation. 
This work is novel but heavily relies on manual paraphrasing and labeling of datasets to find target words.
Moreover, target words are mostly limited to sentiment words. As a result, the model is dominated by these explicit sentiment words and ignores the implied channel.

\subsection{Neural Models}

\citet{DBLP:conf/wassa/GhoshV16} propose a model composed of CNN, LSTM and DNN to detect sarcasm.
As attention mechanism has led to improvements in various NLP tasks, \citet{DBLP:conf/acl/SuTHL18,DBLP:conf/www/XiongZZY19} use attention to capture the relationship of word pairs along with an LSTM to model the entire sentence. 
\citet{DBLP:conf/sigir/LouL0HDX21} design a GCN-based model combining SenticNet~\cite{DBLP:conf/cikm/CambriaLXPK20}, dependency tree and LSTM with GCN~\cite{DBLP:conf/iclr/KipfW17} together, which achieves promising performance. 
Similar to previous studies, to better understand sarcasm, many approaches are able to utilize external information such as emoji expressions~\cite{DBLP:conf/emnlp/FelboMSRL17}, affective knowledge~\cite{DBLP:conf/coling/BabanejadDAP20} and commonsense~\cite{DBLP:journals/taslp/LiPLFW21}. 
\citet{DBLP:journals/csur/JoshiBC17} provide a more comprehensive survey. Moreover, there have been many systems developed for a shared task~\cite{DBLP:conf/acl-figlang/GhoshVM20}.
These models are rarely designed to reflect the essential features of the sarcasm phenomenon.

\section{Dual-Channel Network~(DC-Net)}
\label{sec:model}

The architecture of the proposed DC-Net is shown in Figure~\ref{fig:dual}.
It consists of four modules: \textit{decomposer}, \textit{literal channel}, \textit{implied channel}, and \textit{analyzer}. 
Given an input text, we use the decomposer to split it into two sub-sentences corresponding to the two channels. Then we use these two channels to derive literal and implied representations independently. Lastly, the analyzer predicts whether the text is sarcastic or not by detecting sentiment conflict.

\subsection{Decomposer}

The decomposer module is designed to split input text to the literal and implied channels. 
From numerous sarcastic corpora, we observe that sarcastic texts often contain evident sentiment words. 
More specifically, the literal channel itself is to reflect the intuitive sentiment. So it is reasonable to use sentiment lexicons as direct keywords. The remaining text expresses the implied sentiment.
For example, sentiment words of input text (\eg ``best gift'') represent \textit{positive}, while the remaining  part (\eg ``Final exam is on my birthday'') implies the \textit{negative} sentiment. Shown in Table~\ref{tab:dataset}, proportion of texts that contain sentiment words ranges from 88\% to 96\% in three datasets. 
Hence, using sentiment words to split input well serves the purpose.

\begin{figure}
    \centering
    \includegraphics[scale=0.36]{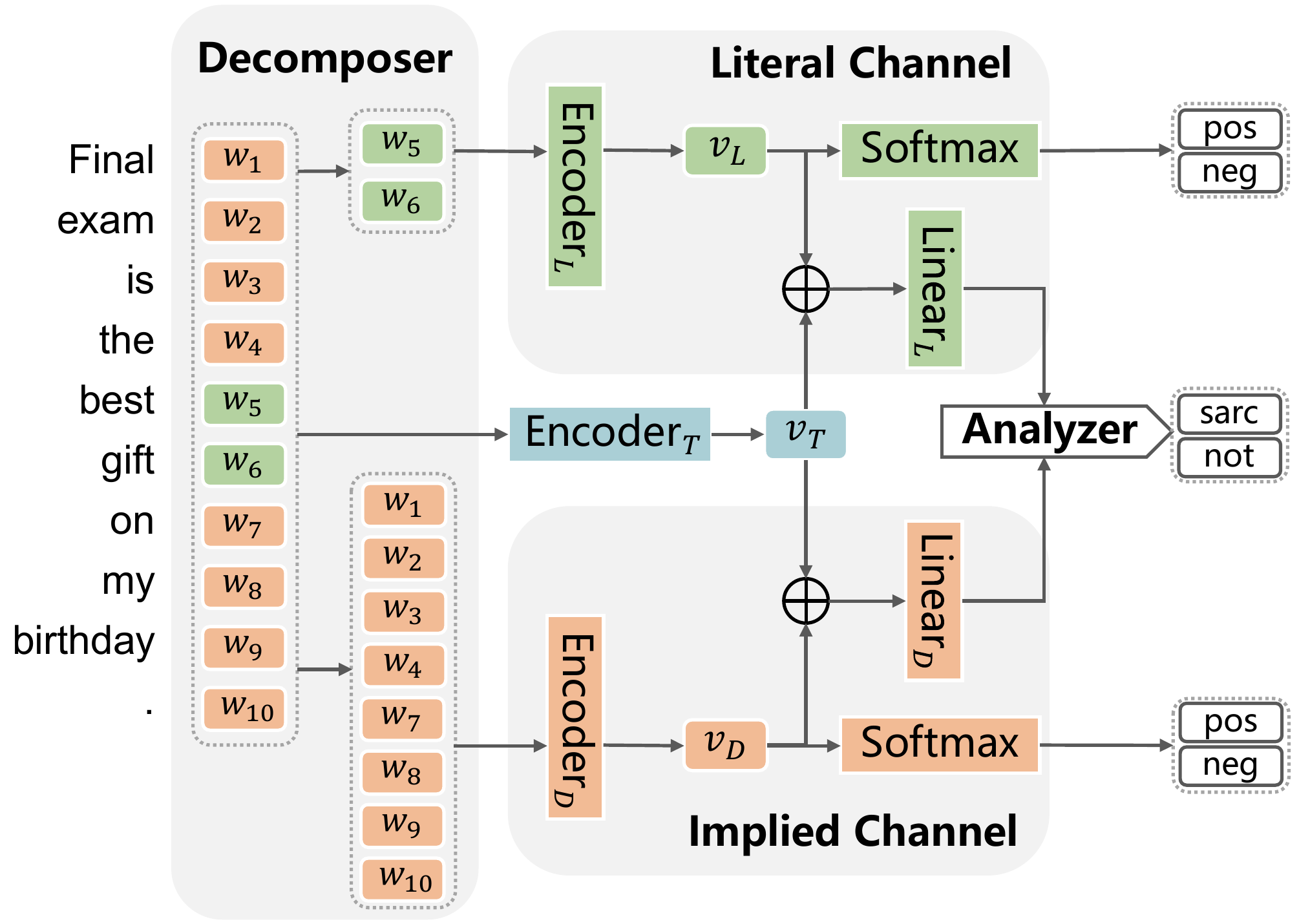}
    \caption{The architecture of the DC-Net.}
    \label{fig:dual}
\end{figure}

Considering a text $W_T=\{w_1, w_2, \dots, w_N\}$ with $N$ words, we decompose it into two pieces: the sentiment words $W_L$, and the remaining text $W_D$ (see Figure~\ref{fig:dual}). $W_L$ is fed to the literal channel, and $W_D$ to the implied channel. 
In this process, we use the sentiment lexicon released in~\citet{wilson2005recognizing} to pick up sentiment words. 
If no sentiment words are matched from the given text, the original text is used as the literal channel's input, which is the same as the implied channel. 
Note that in quite a few texts, sentiment words are adjectives or adverbs, deleting them from sentences has no much impact on the overall semantics. Although the text is not normative as expected after decomposing, we do not fill in the full text with placeholders.

\subsection{Literal Channel}

The literal channel includes an encoder, two linear layers, and a $\softmax$ classifier.
$\text{Encoder}_L$ is used to encode the literal text $W_L$. Then
we can get the literal representation $v_L$ through
\begin{equation}
    v_L=\text{Encoder}_L(W_L).
\end{equation}
Next, we use a $\softmax$ layer to compute the literal sentiment distribution based on the literal representation $v_L$.
\begin{equation}
    \mathcal{P}_l=\softmax(W_r v_L+b_r),
\end{equation}
where, $W_r$ and $b_r$ are parameters of the linear layer.

Considering the semantic complexity of sarcastic texts, a single representation of sentiment words may lose context information. So we adopt another $\text{Encoder}_T$ to encode the original text $W_T$ and obtain the representation of the entire text $v_T$ through
\begin{equation}
    v_T =\text{Encoder}_T(W_T).
\end{equation}

Last, we concatenate the literal state $v_L$ and the entire text's state $v_T$, followed by a linear layer and $\ReLU$ activation function to reduce dimension. Briefly, the final representation $v_L^{'}$ of the literal channel could be formulated as:
\begin{equation}
    v_L^{'} =\ReLU (W_l[v_L;v_T]+b_l),
\end{equation}
where $W_l$ and $b_l$ are parameters of the second linear layer.

\subsection{Implied Channel}

In the implied channel, we also adopt an Encoder with the same structure but different parameters to encode the implied input text $W_D$, and the representation of the implied channel is formulated as:
\begin{equation}
    v_D=\text{Encoder}_D(W_D).
\end{equation}

Similarly, we use $\softmax$ to calculate the implied sentiment distribution based on the implied hidden state $v_D$:
\begin{equation}
    \mathcal{P}_d=\softmax(W_z v_D+b_z),
\end{equation}
where $W_z$ and $b_z$ are parameters.

Again, we concatenate the implied hidden state $v_D$ with the entire text's hidden state $v_T$, followed by a linear layer and an activation layer $\ReLU$. 
The final representation $v_D^{'}$ of the implied channel is formulated as:
\begin{equation}
    v'_D =\text{ReLU}(W_d[v_D; v_T]+b_d),
\end{equation}
where $W_d$ and $b_d$ are parameters.

Note that the structures of the two channels are symmetrical.
However, the two encoders in the two channels do not share parameters, and their inputs are different.  
Since both channels are not specific to particular encoders, the dual-channel framework is able to adapt to mainstream encoders, \eg LSTM~\cite{DBLP:journals/neco/HochreiterS97}, CNN~\cite{DBLP:conf/emnlp/Kim14}, Recursive Neural Network~\cite{DBLP:conf/icml/SocherLNM11}, BERT~\cite{DBLP:conf/naacl/DevlinCLT19}~\etal.
In DC-Net, we adopt Bi-LSTM as encoders for both channels. 

\subsection{Analyzer}
\label{sec:model:ana}

The analyzer is designed to measure the conflict between the literal and the implied channels. 
We concatenate the literal representation $v_L^{'}$ and the implied representation $v_D^{'}$ and feed the result to a $\softmax$ layer.
Other analyzers such as subtraction or cosine similarity also fit our design.
\begin{equation}
    \mathcal{P}_s=\softmax(W_p([v_L^{'},v_D^{'}])+b_p),
\end{equation}
where $W_p$ and $b_p$ are  parameters.

Although sarcasm has a strong correlation to literal sentiment and implied sentiment, we do not have gold labels for both sentiments. Hence, requesting the model to directly output sentiments on both channels may confuse the model. 
For this reason, we develop the objective function of sarcasm classification by adding objectives of the literal and implied channels.

\subsection{Training Objective}

The training objective of the proposed DC-Net model considers three aspects. One is to minimize the cross-entropy loss of the sarcasm probability distribution. The other two are to minimize the cross-entropy losses of the literal and that of the implied sentiment probability distributions respectively. 

\paratitle{Sarcasm Objective.}
The sarcasm objective is to ensure the basic ability of detection. Hence, we use cross-entropy loss of sarcasm classification. The objective $J_s$ is formulated as:
\begin{equation}
    J_s(\theta)=\sum \crossentropy(y_s, \mathcal{P}_s),
\end{equation}
where $\mathcal{P}_s$ denotes the sarcasm probability distribution of the text. The groundtruth of the sarcasm label is $y_s$.

\paratitle{Literal Sentiment Objective.}
Due to the expensive manual annotations, we use sentiment words for approximate labeling, which is widely used in~\citet{DBLP:conf/aaai/Eisenstein17,DBLP:journals/coling/TaboadaBTVS11,DBLP:conf/kdd/HuL04}.
In our implementation, we determine the literal sentiment label based on the number of words with positive sentiment and the words with negative sentiment in input text. 
For sarcastic texts, if the number of positive words is greater than that of negative words, the literal sentiment label is positive and the implied sentiment label is negative, and vice versa. 
For non-sarcastic texts, both the literal sentiment label and the implied sentiment label are the same, determined by the number of positive/negative sentiment words.

The literal sentiment classification objective is then formulated as:
\begin{equation}
   J_l(\theta)=\sum \crossentropy(y_l,\mathcal{P}_l),
\end{equation}
where $\mathcal{P}_l$ is the literal sentiment probability distribution. 
The label generated by the labeling processing of the literal sentiment is $y_l$.

\paratitle{Implied Sentiment Objective.}
We observe that literal sentiment and implied sentiment of sarcastic texts are often opposite. 
Using the implied labels based on the automatic labeling processing, we calculate the implied sentiment classification objective by
\begin{equation}
   J_d(\theta)=\sum \crossentropy(y_d, \mathcal{P}_d),
\end{equation}
where $\mathcal{P}_d$ denotes the implied sentiment probability distribution. 
The label generated by the labeling processing of the implied sentiment is $y_d$.

Considering these three objectives, we obtain the final objective function $L$ by adding them together:
\begin{equation}
    L(\theta)=\lambda_1 J_s(\theta)+\lambda_2 J_l(\theta)+\lambda_3 J_d(\theta),
\end{equation}
where $\theta$ is the parameter set of the model. 
$\lambda_1$, $\lambda_2$ and $\lambda_3$ are used to leverage the contributions of the three objectives.

\section{Experiment}
\label{sec:experiment}

\subsection{Datasets and Implementation Details}

We conduct experiments on three benchmark datasets: IAC-V1, IAC-V2, and Tweets. These datasets do not contain context information such as historical tweet posts and user profiles. All of them have been widely used in evaluating sarcasm detection. 

\begin{itemize}
\item \textbf{IAC-V1} is collected from online political debates forum\footnote{http://www.4forums.com/political/}. It is the subset of the Internet Argument Corpus~\cite{lukin-walker-2013-really}. 
The written language of IACs is English. Each instance, typically a sentence, is annotated with  sarcasm label, either ``sarcasm'' or ``non-sarcasm''. 
Compared to tweets, texts of IAC are much longer and more normative. 
\item \textbf{IAC-V2}~\cite{DBLP:conf/sigdial/OrabyHRHRW16} contains more data than IAV-V1~(the two versions have a few overlaps). IAC-V2 divides sarcasm into three sub-types, (\ie general sarcasm, hyperbole, and rhetorical questions). We use the largest subset~(general sarcasm) in our experiments.
\item \textbf{Tweets} dataset written in English is proposed in SemEval 2018 Task 3 Subtask A~\cite{DBLP:conf/semeval/HeeLH18}. Each instance (\ie a sentence) is labeled sarcastic or non-sarcastic. There are three variations of the text in this dataset: (i) original texts, (ii) texts with hashtags removed, and (iii) texts with hashtags and emoji expressions removed.
Hashtags like "\#not", "\#sarcasm", and "\#irony", are originally obtained from users. The hashtags are also used as prior knowledge for collecting sarcastic posts. In our experiments, we used the version without hashtags.
\end{itemize}

Table~\ref{tab:dataset} reports the statistics. We observe that more than $88\%$ of the texts contain sentiment word(s). Hence, it is reasonable to decompose the original text into sentiment words and non-sentiment words, as inputs to the literal channel and implied channel, respectively.
The number of instances in the three datasets is between $1k$ and $6k$. All three datasets are class-balanced. The ratio of sarcastic instances and non-sarcastic instances is nearly $\text{1:1}$.
Due to the small size, the split of train/valid/test is important to avoid over-fitting. 
For Tweets dataset, we follow the official train/test split. Then we randomly select $5\%$ from training as valid sub-dataset.
There is no official train/valid/test split for the two IAC datasets, so we split IAC datasets following the same ratio of Tweets. The baselines papers do not provide the split (or not conduct experiments on IAC datasets). So we cannot directly adopt the results of baselines reported in their original papers. Hence, we re-implement all baseline models on IAC-V1 and IAC-V2 datasets.

\begin{table}
\caption{Statistics of datasets. Avg $\ell$ denotes the average length of texts in the number of tokens. $s$ ratio is the proportion of texts that contain sentiment words.}
\scalebox{0.9}{
  \centering
    \begin{tabular}{l|rrrrr}
    \toprule
    Dataset & Train & Valid & Test &  Avg $\ell$ & $s$ ratio \\
    \midrule
    IAC-V1\tablefootnote{https://nlds.soe.ucsc.edu/sarcasm1} & 1,596  & 80    & 320   & 68    & 91\% \\
    IAC-V2\tablefootnote{https://nlds.soe.ucsc.edu/sarcasm2} & 5,216  & 262   & 1,042  & 43    & 96\% \\
    Tweets\tablefootnote{https://github.com/Cyvhee/SemEval2018-Task3} & 3,634  & 200   & 784   & 14    & 88\% \\
    \bottomrule
    \end{tabular}
    }
  \label{tab:dataset}
\end{table}

There are another three datasets for sarcasm detection.  \citet{riloff2013sarcasm} and~\citet{ptavcek2014sarcasm} propose another two datasets based on Tweets, but they only provide tweet IDs. Due to modified authorization status, lots of tweets are unavailable or deleted. For this reason, we could not experiment on these two Tweet datasets. 
\citet{DBLP:conf/lrec/KhodakSV18} build a large self-annotated dataset from the Reddit forum platform. 
This dataset contains rich context information including posts, comments, responses, and authors. 
Since our work focuses on text-based sarcasm recognition, we do not use this dataset.

\begin{table*}[thbp]
\caption{The precision, recall, and macro $F1$ of sarcasm recognition. The results marked with * are from \citet{DBLP:conf/semeval/HeeLH18}. The best results are in boldface and second-best underlined.}
\scalebox{0.9}{
  \centering
    \begin{tabular}{l|cccc|cccc|cccc}
    \toprule
    \multirow{2}{*}{Model}
           & \multicolumn{4}{c|}{IAC-V1} & \multicolumn{4}{c|}{IAC-V2} & \multicolumn{4}{c}{Tweets} \\
    \cline{2-13}
     & Pre.  & Rec.  & F1    & Acc.  & Pre.  & Rec.  & F1    & Acc.  & Pre.  & Rec.  & F1    & Acc. \\
    \midrule
    UCDCC & 58.6  & 58.6  & 58.5  & 58.5 & 67.1  & 67.0  & 67.0  & 67.0   & \textbf{78.8$^*$}  & 66.9$^*$  & 72.4$^*$  & \textbf{79.7$^*$}   \\
    THU-NGN & 64.4  & 64.3  & 64.2  & 64.3  & 73.3  & 73.3  & 73.3  & 73.3 & 63.0$^*$  & \textbf{80.1$^*$}  & 70.5$^*$  & 73.5$^*$   \\
    Bi-LSTM & 64.6 & 64.6 & 64.6 & 64.6 & 79.8 & 79.7 & 79.7 & 79.7 & 71.8 & 71.7 & 71.7 & 73.0  \\ 
    AT-LSTM & \underline{65.9}  & \underline{65.5}  & \underline{65.3}  & \underline{65.5} & 76.7  & 76.2  & 76.1  & 76.2 & 70.8  & 71.6  & 70.0  & 70.2   \\
    CNN-LSTM-DNN & 61.5  & 61.2  & 60.9  & 61.1  & 75.4  & 75.3  & 75.2  & 75.3   & 71.9  & 72.9  & 71.9  & 72.3 \\
    MIARN  & 65.6  & 65.2  & 64.9  & 65.2 & 75.4  & 75.3  & 75.2  & 75.3  & 68.6  & 68.8  & 68.8  & 70.2  \\
    ADGCN & 64.3 & 64.3 & 64.3 & 64.3 & \underline{81.0} & \underline{80.9} & \underline{80.9} & \underline{80.9} & 72.6 & 73.2 & \underline{72.8} & 73.6 \\ 
    \midrule
    DC-Net & \textbf{66.6} & \textbf{66.5} & \textbf{66.4} & \textbf{66.5} & \textbf{82.2} & \textbf{82.1} & \textbf{82.1} & \textbf{82.1} & \underline{76.4} & \underline{77.5} & \textbf{76.3} & \underline{76.7} \\
    \bottomrule
    \end{tabular}
    }
  \label{tab:main}
\end{table*}

\paratitle{Implementation Details.}
We use $300$-dimensional Glove~\cite{pennington2014glove} embeddings to initialize word vectors.
There is a checkpoint every $16$ mini-batch, and the batch size is $32$. 
For Tweets dataset, the dropout on embeddings is set to $0$, while for IAC datasets it is set to $0.5$.
Adam~\cite{DBLP:journals/corr/KingmaB14} is used to optimize our model. The parameters $\beta_1$ and $\beta_2$ of Adam are set to $0.9$ and $0.999$.
The learning rates for model parameters except word vectors are $\text{1e-3}$, and $\text{1e-4}$ for word vectors.  Our model is implemented with Pytorch\footnote{https://pytorch.org}~(version 1.7.0).

On IAC datasets, all of the loss contributions $\lambda_1, \lambda_2, \lambda_3$ of our DC-Net model are set to $1$. On Tweets, they are set to $1$, $\text{1e-4}$, and $\text{3e-1}$, respectively. The hyperparameters are searched over the validation sub-dataset.

\subsection{Compared Methods}

We evaluate our model against the following baselines:

\paratitle{UCDCC}~\cite{DBLP:conf/semeval/GhoshV18} is a siamese LSTM model exploiting Glove word embedding features.  The method designs a lot of rules to preprocess Twitter data.
It achieves the best performance on SemEval 2018 Task 3 Subtask A.

\paratitle{THU-NGN}~\cite{wu2018thu_ngn} consists of densely connected LSTMs based on word embeddings, sentiment features, and syntactic features. It ranks second on SemEval 2018 Task 3 Subtask A.

\paratitle{Bi-LSTM}~\cite{DBLP:journals/neco/HochreiterS97} is a variant of RNN, which could learn long-term dependencies and bidirectional information. 
    
\paratitle{AT-LSTM}~\cite{wang2016attention} is an LSTM model followed by a neural attention mechanism. It could attend the important part of the input.

\paratitle{CNN-LSTM-DNN}~\cite{DBLP:conf/wassa/GhoshV16} is a combination of CNN, LSTM, and DNN. It stacks two layers of convolution and two LSTM layers, then passes the output to a DNN for prediction.

\paratitle{MIARN}~\cite{DBLP:conf/acl/SuTHL18} learns the intra-sentence relationships of word pairs and the sequential relationships of a given text.

\paratitle{ADGCN}~\cite{DBLP:conf/sigir/LouL0HDX21} is a GCN-based method with sentic graph and dependency graph\footnote{We employ spaCy toolkit to derive dependency tree.}. The initial input of GCN is the hidden state of Bi-LSTM.

\subsection{Main Experiment Results}

Table~\ref{tab:main} shows that our DC-Net achieves the best macro $F1$ results across all datasets.
On Tweets dataset, DC-Net achieves about $3.5\%$ improvement in $F1$ score than the best baseline. On IAC-V2 dataset, our model outperforms the second-best by $1.2\%$ in $F1$. 
Surprisingly, compared with the basic encoder model Bi-LSTM, our DC-Net boosts the performance up to $5\%$ and $3\%$ respectively on Tweets and IAC-V2, demonstrating the effectiveness of our dual-channel design. 
For Tweets dataset, the average length of texts is 14 words, which leads to a lack of information for sarcasm recognition. 
Nevertheless, our DC-Net improve $3.5\%$ on $F1$ compared with the previous state-of-the-art ADGCN.

Interestingly, UCDCC achieves the best precision of $78.8\%$ and accuracy of $79.7\%$ on Tweets dataset. Besides, THU-NGN gets the best recall at $80.1\%$ on Tweets. This is because UCDCC designs targeted rules to preprocess the input text and it achieves the best performance on SemEval 2018 Task 3 Subtask A. Rules could improve precision effectively, but they are hard to take recall into account at the same time. So the $F1$ is not good enough. The last place performance of UCDCC on IAC-V1/V2 also supports this point. These designed rules are hard to fit missing instances and other domains. Similarly, THU-NGN uses linguistic knowledge such as sentiment and syntactic, so it achieves the highest recall on Tweets but it cannot perform equally well on other datasets. That is, rules have limitations in handling this task.

The previous state-of-the-art ADGCN  achieves second-best on IAC-V2 and Tweets. However, on IAC-V1 dataset, ADGCN performs not as well as the result reported in their paper. IAC-V1 dataset is relatively small so the train/valid/test split has a significant impact.
Our experiments also show that MIARN's performance is not as good as expected. This indicates that the basic utilization of word pair correlation is not enough to improve the performance of sarcasm detection. Bi-LSTM, AT-LSTM, and CNN-LSTM-DNN methods are all based on LSTM. Thus the performances of these models on Tweets and IAC-V1 are close.

\subsection{Comparison with BERT}

BERT has contributed to significant improvements on various NLP tasks. To do a comprehensive comparison, we apply the dual-channel framework to BERT~\cite{DBLP:conf/naacl/DevlinCLT19} model by using BERT as the encoder. The new model with BERT is named DC-Net~(w/ BERT).
Table~\ref{tab:bert} reports the experimental results.

As expected, the DC-Net~(w/ BERT) model achieves significant improvement compared with the basic BERT. This result shows that our dual-channel framework is adaptable and effective. Interestingly, we observe that BERT-based methods perform not well enough compared with its huge improvement on other NLP tasks.
This can be attributed to the fact that the corpus of pre-trained BERT contains more deterministic data~(\eg only one meaning without sentiment conflict). However, sarcasm is a niche linguistic phenomenon.  The poor performance of BERT further reinforces that sarcasm recognition is a difficult task. It tells us that applying well-performed classification methods directly is difficult to achieve desirable performance.

\begin{table}
\caption{The precision, recall, and macro $F1$ of models including BERT, DC-Net with BERT as Encoder, and DC-Net with Bi-LSTM as Encoder.}
\scalebox{0.9}{
  \centering
    \begin{tabular}{l|cccc}
    \toprule
    \multirow{2}{*}{Model}
          & \multicolumn{4}{c}{Tweets} \\
    \cline{2-5}
     & Pre.  & Rec.  & F1    & Acc. \\
    \midrule
    BERT  & 69.1  & 67.6  & 68.1  & 71.6  \\
    DC-Net~(w/ BERT) & 70.2 & 70.7 & 70.4 & 71.3 \\
    DC-Net~(w/ Bi-LSTM) & \textbf{76.4}  & \textbf{77.5}  & \textbf{76.3}  & \textbf{76.7}  \\ 
    \bottomrule
    \end{tabular}
    }
  \label{tab:bert}
\end{table}

\subsection{Ablation Study}

Recall that the model training (see Section~\ref{sec:model}) contains three objectives: sarcasm recognition, literal sentiment classification, and implied sentiment classification.
To study the effect of the three objectives, we conduct ablation study on Tweets dataset. 

\begin{table}
    \caption{Ablation study on Tweets dataset. $J_s$ denotes using sarcasm loss only. $J_s$+$J_d$ means using sarcasm and implied loss. $J_s$+$J_l$ means using sarcasm and literal loss. $J_s$+$J_l$+$J_d$ denotes using sarcasm, literal, and implied loss.}
  \centering
  
    \begin{tabular}{l|cccc}
    \toprule
    \multirow{2}{*}{Objective}
          & \multicolumn{4}{c}{Tweets} \\
    \cline{2-5}
      & Pre.  & Rec.  & F1    & Acc. \\
    \midrule
    $J_s$ & 74.6 & 75.4 & 74.8 & 75.4 \\
    $J_s$+$J_d$ & 74.2 & 75.2 & 74.0 & 74.4 \\
    $J_s$+$J_l$ & 73.0 & 74.0 & 72.8 & 73.1 \\
    $J_s$+$J_l$+$J_d$ & \textbf{76.4}  & \textbf{77.5}  & \textbf{76.3}  & \textbf{76.7} \\
    \bottomrule
    \end{tabular}
  \label{tab:loss}
\end{table}

Table~\ref{tab:loss} lists the result of ablation study. 
As expected, the model with both literal and implied losses performs the best. 
Interestingly, the model using sarcasm recognition loss with single channel loss (\ie literal and implied) performs worse than the model using only sarcasm recognition loss. This is because adding literal and implied sentiment classification objectives interferes with the judgment of the model.
By adding both literal and implied sentiment classification losses, the model's performance improves $1.5$ points in $F1$ score. 
This is very important because it reveals that the dual channels are effective. There is no effect or the opposite effect when single channel is applied alone. However, once dual-channel is used, the performance improves largely.
It reveals that the dual channels complement each other. Conflict detection could recognize sarcasm when both of them are considered.

\subsection{Effectiveness of DC-Net by Visualization}

To verify the rationality and effectiveness of our proposed DC-Net, we adopt t-SNE~\cite{van2008visualizing} to visualize high-dimensional vector representations based on the test sub-dataset of IAC-V2 (with largest data).

\begin{figure}
	\centering
	\subfigure[Literal and implied reps. of each channel]{
		\hspace{-2mm}\begin{minipage}[b]{0.4\linewidth}
			\includegraphics[width=1.1\linewidth]{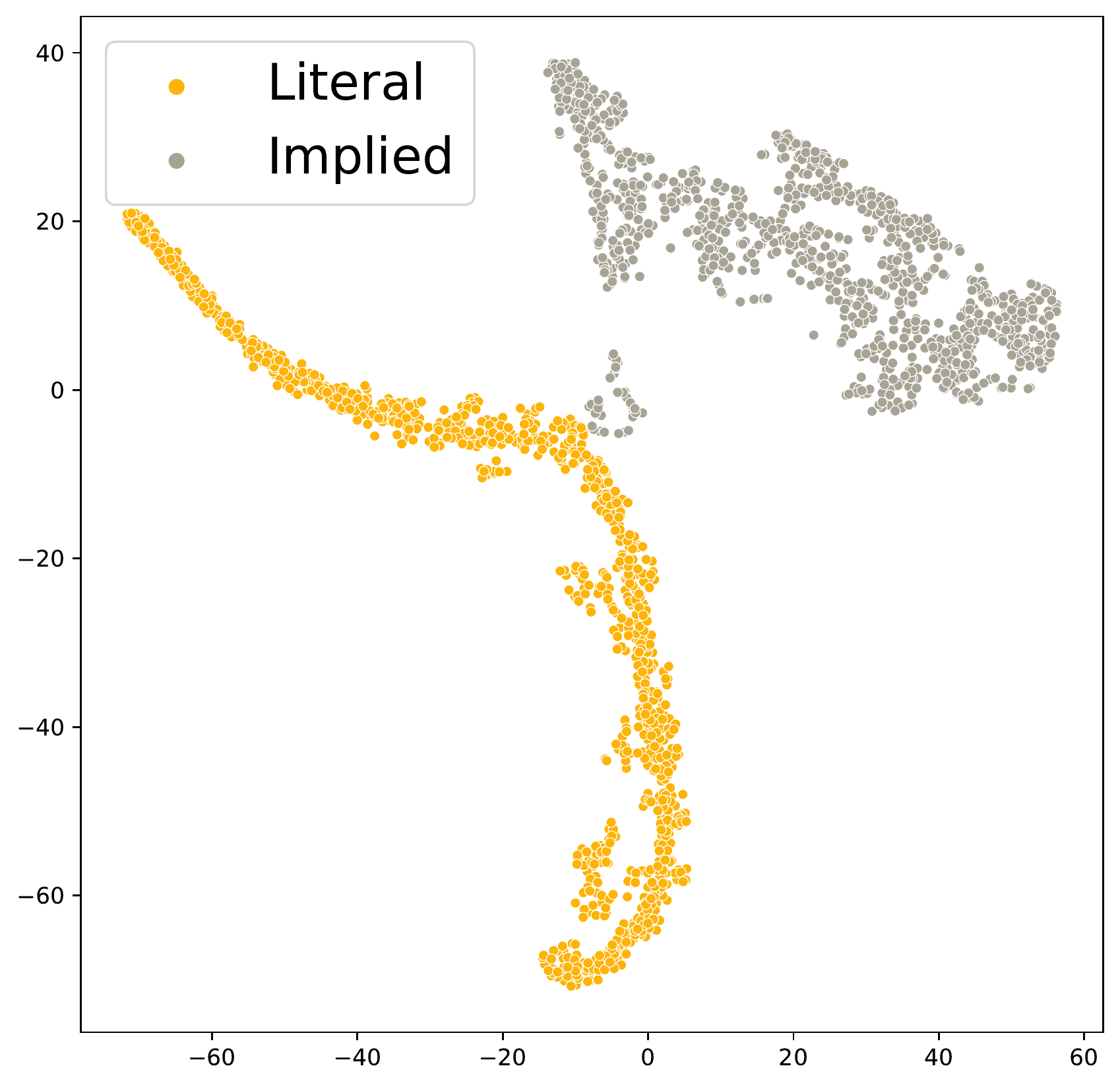}
		\end{minipage}
		\label{fig:decompose}
	}\quad
    	\hspace{2.6mm}\subfigure[Sarc. and non-sarc. reps. in analyzer module]{
    		\begin{minipage}[b]{0.4\linewidth}
   		 	\includegraphics[width=1.1\linewidth]{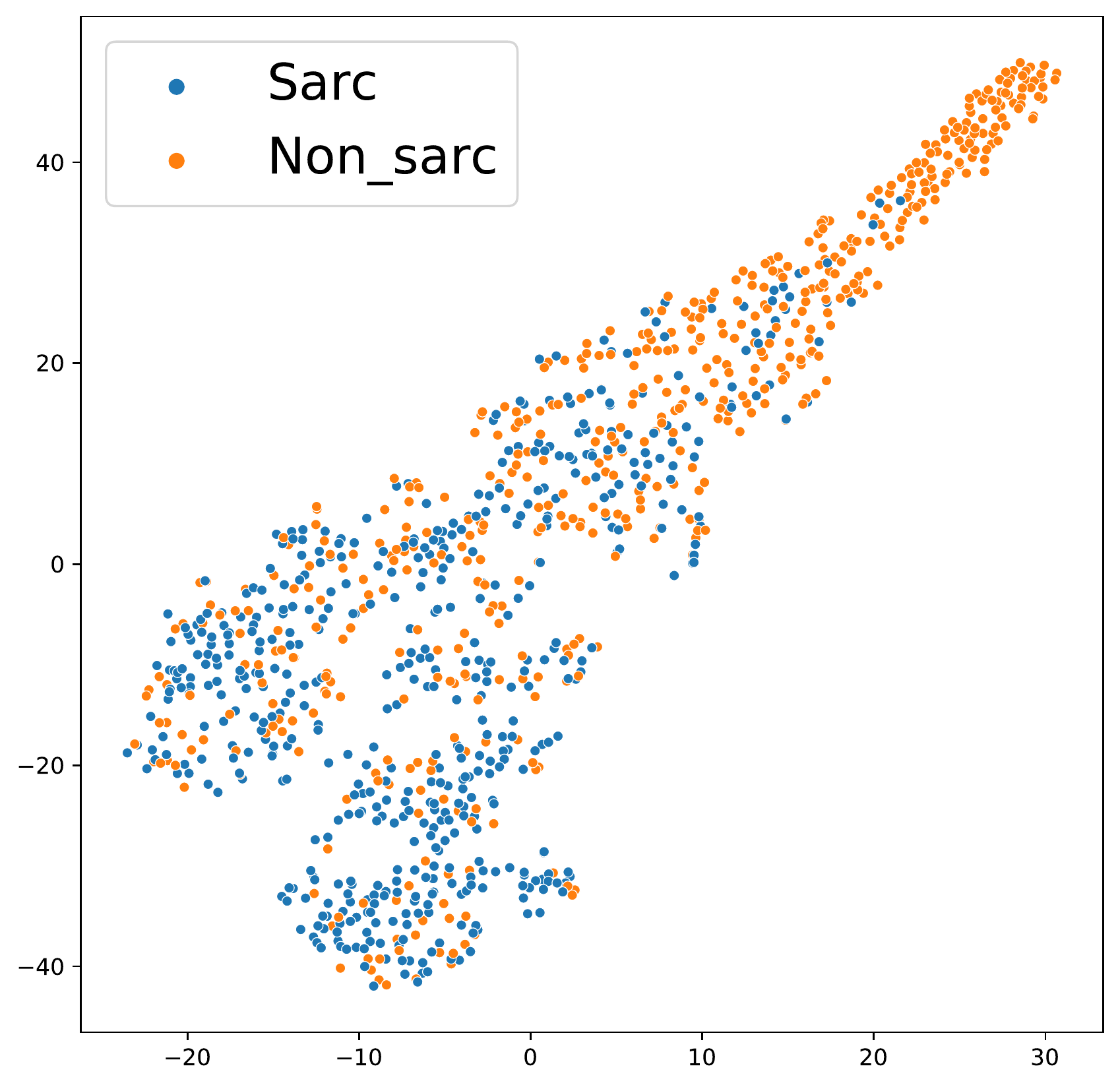}
    		\end{minipage}
		\label{fig:sen}
    	}
	\caption{Results of t-SNE visualization}
	\label{fig:t-sne}
\end{figure}

To figure out the effect of each channel, we visualize the representations of the literal channel and the implied channel. 
Figure~\ref{fig:decompose} shows the visualization of literal representation $v_L^{'}$ and implied representation $v_D^{'}$. Recall that the decomposer module splits the original text into sentiment words and the remaining. We observe that there is a clear separation between literal and implied representations from Figure~\ref{fig:decompose}. This strongly indicates that our dual-channel framework is capable of effectively separating the representations of the two channels.

To get into the essence of sarcastic and non-sarcastic texts, we visualize the sarcastic and non-sarcastic representations.
Figure~\ref{fig:sen} shows the sarcastic and non-sarcastic representations in the analyzer module.
We observe that non-sarcastic texts focus on the upper right corner, while sarcastic texts scatter on the lower left corner. 
It reveals that the sarcasm patterns are complex and changeable. Nevertheless, the dividing line between the two is relatively clear.
To this end, explicitly separating the literal and implied channels is necessary and effective. 
Further, DC-Net makes a distinct difference between sarcastic and non-sarcastic representations, which greatly promotes the performance of the dual-channel framework.
We also plot the sarcastic representation and non-sarcastic representation of each channel respectively, which show the same trend as Figure~\ref{fig:sen}. So they are not detailed here.

\subsection{Flexibility of Dual-Channel Framework} 

\paratitle{Flexibility of encoder.}
The dual-channel framework is flexible and generic, and can be realized by plugging in existing sarcasm recognition models, \eg MIARN, or classification models, \eg AT-LSTM, Bi-LSTM, and BERT. 
Therefore, we use these methods as the encoder to examine the flexibility of our proposed framework. The changing range on macro $F1$ from original baseline models to dual-channel models is shown in Table~\ref{tab:flex}.

As expected, the performance of baseline models has different degrees of improvement on all datasets after applying dual-channel framework. For relatively simple models such as MIARN and Bi-LSTM, the improvement could be up to $4.8\%$. 
Interestingly, for complex models like BERT, the improvement is up to $2.3\%$. As we mentioned earlier, the basic BERT performs not good enough because sarcasm is a niche language phenomenon and the training dataset of BERT contains few sarcasm texts. 
After applying the dual-channel framework to BERT, the performance for sarcasm recognition improves a lot. 
These indicate that our designed framework is able to fit various encoders with a significant improvement.

\begin{table}
    \caption{The macro $F1$ changes from basic models to dual-channel based models.}
  \centering
    \begin{tabular}{l|c|c|c}
    \toprule
    \multirow{2}[3]{*}{Basic Model} & \multicolumn{3}{c}{Changing Range on F1} \\
\cmidrule{2-4}          & IAC-V1 & IAC-V2 & Tweets \\
    \midrule
    AT-LSTM  & ↑ 0.4 & ↑ 1.1 & ↑ 1.5  \\
    BERT   & ↑ 0.4 & ↑ 1.7 & ↑ 2.3 \\
    MIARN & ↑ 1.1  & ↑ 2.8  & ↑ 4.8 \\
    Bi-LSTM & ↑ 1.8 & ↑ 2.4 & ↑ 4.6 \\
    \bottomrule
    \end{tabular}
  \label{tab:flex}
\end{table}

\paratitle{Flexibility of analyzer.}
The analyzer module is used to measure the difference between the literal channel and the implied channel. 
As we described in Section~\ref{sec:model:ana}, other analyzer methods such as  concatenation and subtraction could be applicable. To this end, we compare different analyzer methods. Table~\ref{tab:analyzer} shows the results. We observe that concatenation performs better than subtraction on all datasets. It is because concatenation holds more useful information and DC-Net could compare the difference between the two input representations. However, subtraction only outputs the margin between the two representations. It loses the original values which also contain useful information.

\begin{table}
    \caption{Comparisons of different analyzer methods.}
  \centering
    \begin{tabular}{l|c|c|c}
    \toprule
    \multirow{2}[4]{*}{Analyzer} & \multicolumn{3}{c}{F1} \\
\cmidrule{2-4}          & IAC-V1 & IAC-V2 & Tweets \\
    \midrule
    Subtraction & 65.1  & 80.7  & 75.2 \\
    Concatenation & \textbf{66.4}  & \textbf{82.1}  & \textbf{76.3} \\
    \bottomrule
    \end{tabular}
  \label{tab:analyzer}
\end{table}

\section{Conclusion}
\label{sec:conclusion}

In this study, we argue that the essential characteristic of sarcastic text is the conflict between literal and implied sentiments in the same sentence. 
To this end, we propose a dual-channel framework to recognize sarcasm by decomposing the input text into the literal channel and the implied channel.
Based on this framework, we develop DC-Net. DC-Net is capable of exploiting the literal sentiment by encoding the sentiment words of input text, and exploiting the implied sentiment by encoding the remaining text. Experiments show that the proposed DC-Net achieves state-of-the-art performance.

\section{Limitation}
\label{sec:limitation}

Sarcasm as a complex linguistic phenomenon has various patterns, \eg \textit{text with word/phrase pair sentiment conflict}. Nevertheless, sentiment conflicts are common in sarcasm texts. In this paper, we make the very first attempt to recognize sarcasm by detecting sentiment conflict.
More importantly, our proposed dual-channel framework could be further developed to detect more sentiment conflict patterns.
For now, we use sentiment words as a static decomposer. This intuitive method can cover common sarcasm patterns but not all. Therefore, how to minimize the dependence on sentiment words is an important research direction.

Another limitation is that we assume that sentiment polarity is decided by the sentiment lexicon approximately in the analyzer module. While the assumption is widely accepted, there is still a gap between approximate label and groundtruth.
In the current design, we adopt a soft weighting mechanism to detect sentiment conflict between the two channels. 
We expect that the model could output the opposite sentiment labels directly, which is a more effective way to express conflict.

\section*{Acknowledgments}
This work is supported by the National Science Foundation of China (NSFC No. 62106249, No. 61902382 and No. 61972381), the Youth Innovation Promotion Association CAS under Grants No. 20144310, the Lenovo-CAS Joint Lab Youth Scientist Project, and the Foundation and Frontier Research Key Program of Chongqing Science and Technology Commission (No. cstc2017jcyjBX0059).

\bibliography{anthology,custom}
\bibliographystyle{acl_natbib}

\end{document}